\documentclass{article} % For LaTeX2e
\usepackage[final]{colm2026_conference}

\usepackage{microtype}
\usepackage{hyperref}
\usepackage{url}
\usepackage{booktabs}
\usepackage{amsmath}
\usepackage{amssymb}
\usepackage{graphicx}
\usepackage{subcaption}
\usepackage{xcolor}
\usepackage{multirow}
\usepackage{comment}
\usepackage{algorithm}
\usepackage{algorithmic}
\usepackage{lineno}
\usepackage{wrapfig}

\definecolor{darkblue}{rgb}{0, 0, 0.5}
\hypersetup{colorlinks=true, citecolor=darkblue, linkcolor=darkblue, urlcolor=darkblue}

\title{HISA: Efficient Hierarchical Indexing \\ for Fine-Grained Sparse Attention}

\author{Yufei Xu\thanks{Equal contribution: Yufei Xu and Fanxu Meng.},~Fanxu Meng\footnotemark[1]\hspace{0.5em}\thanks{Project leader: Fanxu Meng (\texttt{mengfanxu@xiaohongshu.com}).},~Fan Jiang,~Yuxuan Wang,~Ruijie Zhou,~Zhaohui Wang,\\
  \textbf{Jiexi Wu,~Zhixin Pan,~Xiaojuan Tang,~Wenjie Pei,~Tongxuan Liu,~Di Yin,} \\
  \textbf{Xing Sun,~Muhan Zhang\thanks{Corresponding author: Muhan Zhang (\texttt{muhan@pku.edu.cn}).}}\\
  \centering\href{https://github.com/MuLabPKU/TransArch}{https://github.com/MuLabPKU/TransArch}}

\begin{document}

\ifcolmsubmission
\linenumbers
\fi

\maketitle
% ===========================================================================
% ABSTRACT
% ===========================================================================
\begin{abstract}
Token-level sparse attention mechanisms, exemplified by DeepSeek Sparse Attention (DSA), achieve fine-grained key selection by scoring every historical key for each query through a lightweight indexer, then computing attention only on the selected subset.
While the downstream sparse attention itself scales favorably, the indexer must still scan the entire prefix for every query, introducing an $\mathcal{O}(L^2)$ per-layer bottleneck that grows prohibitively with context length.
We propose \textbf{HISA} (\textbf{H}ierarchical \textbf{I}ndexed \textbf{S}parse \textbf{A}ttention), a plug-and-play replacement for the indexer that rewrites the search path from a flat token scan into a two-stage hierarchical procedure: (1) a block-level coarse filtering stage that scores pooled block representations to discard irrelevant regions, followed by (2) a token-level refinement stage that applies the original indexer exclusively within the retained candidate blocks.
HISA preserves the token-level top-$k$ index interface consumed by the downstream Sparse MLA operator and requires \textbf{no additional training}.
In end-to-end serving with DeepSeek-V3.2, HISA improves throughput by $1.65\times$ and time to first token by $1.70\times$ over the official DSA implementation, despite the HISA-specific operators being implemented in TileLang/Triton while the official DSA baseline uses highly optimized hand-tuned CUDA kernels.
With matched TileLang implementations of both indexers, controlled microbenchmarks show up to \textbf{$3.75\times$ indexer speedup} at 64K context.
On Needle-in-a-Haystack and LongBench, we directly replace the indexer in DeepSeek-V3.2 and GLM-5 with our HISA indexer, without any finetuning. 
HISA closely matches the original DSA in quality, while substantially outperforming block-sparse baselines.
\end{abstract}

% ===========================================================================
% 1  INTRODUCTION
% ===========================================================================
\section{Introduction}
\label{sec:intro}

Serving large language models (LLMs) \citep{gpt54, claude46, gemini3, llama4, qwen35, dsviii, minimax01, kimik2} over long contexts remains a central systems challenge. As context windows grow from 128K to 1M tokens and beyond---driven by demands for agentic multi-turn reasoning, long-document understanding, and native multimodal processing---the quadratic cost of self-attention becomes a dominant bottleneck in both prefill latency and memory consumption~\citep{dao2022flashattention, dao2023flashattention2}.

A productive line of work tackles this challenge through \emph{sparse attention}: instead of attending to all key--value pairs, each query selects a small subset of the most relevant tokens and computes attention only over that subset. DeepSeek-V3.2~\citep{deepseekv32} adopts a \emph{token-level} sparse attention paradigm, in which a lightweight \emph{indexer} scores every historical token for each query, selects the top-$k$ highest-scoring keys, and forwards only those keys to a downstream Sparse Multi-Head Latent Attention (Sparse MLA). This design has also been adopted in GLM-5~\citep{glm5} and provides strictly finer-grained selection than block-level methods such as MoBA~\citep{lu2025moba} and Native Sparse Attention~\citep{yuan2025native}.

However, the token-level sparse paradigm introduces a subtler bottleneck. Although the downstream attention is sparse and cheap, the indexer itself must score every token in the prefix for every query. Concretely, if the prefix length is $L$ and the indexer runs once per query per layer, the per-layer indexing cost is $\mathcal{O}(L^2)$---the same asymptotic scaling as dense attention. As context lengths push toward 128K or 1M tokens, the indexer can transition from a negligible overhead into the dominant cost component.

This observation motivates a natural question: \emph{can we reduce the indexer's search cost without changing the final sparse attention pattern it produces?} In other words, can we rewrite the search \emph{path} while preserving the search \emph{result}?

\begin{figure}[t]
\centering
\begin{subfigure}[t]{0.48\textwidth}
    \centering
    \includegraphics[width=\linewidth]{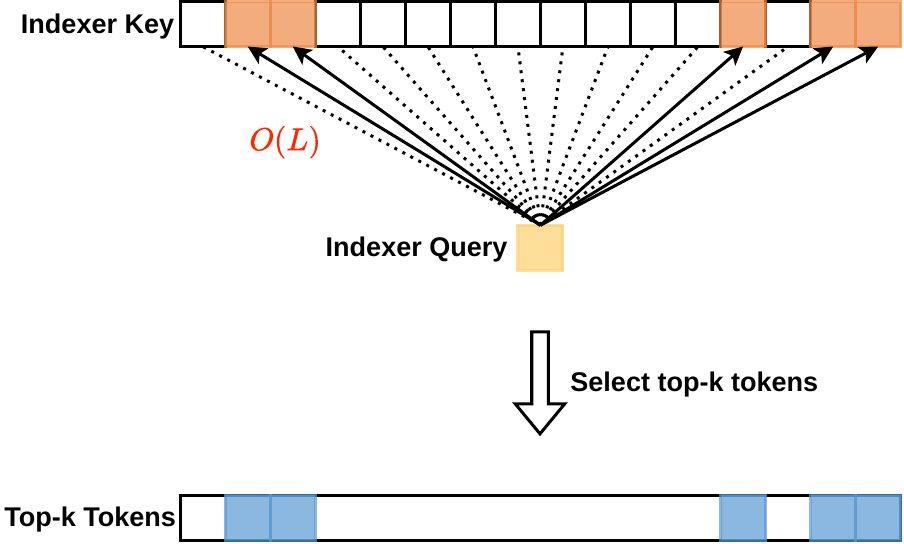}
    \caption{Original DSA: token-wise indexer.}
    \label{fig:token_indexer}
\end{subfigure}
\hfill
\begin{subfigure}[t]{0.48\textwidth}
    \centering
    \includegraphics[width=\linewidth]{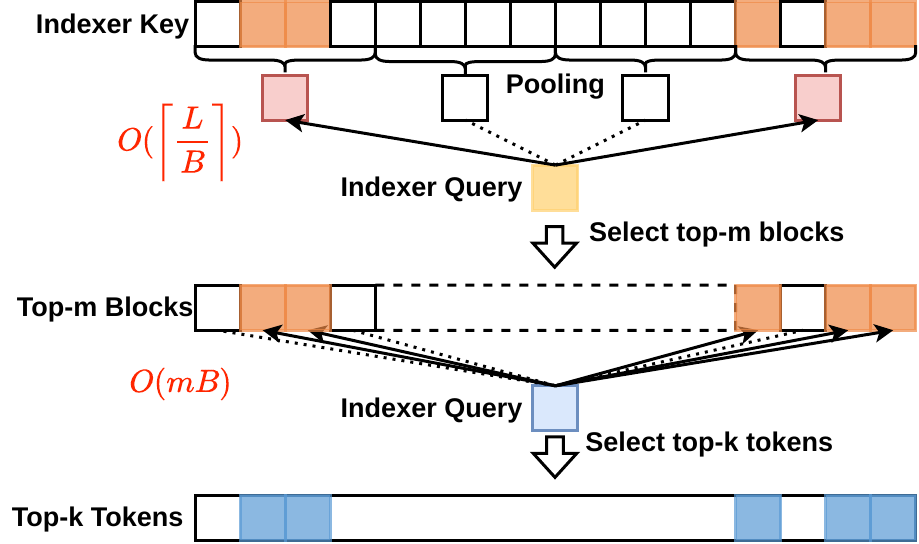}
    \caption{Our HISA: block-to-token indexer.}
    \label{fig:hierarchical_indexer}
\end{subfigure}
\caption{Comparison of the DSA token-wise indexer (left) and our HISA hierarchical block-level coarse filter followed by token-level refinement (right). Both produce the same data structure---a per-query set of $k$ token indices---consumed by the downstream Sparse MLA operator.}
\label{fig:method}
\end{figure}

We answer affirmatively with \textbf{HISA} (\textbf{H}ierarchical \textbf{I}ndexed \textbf{S}parse \textbf{A}ttention). HISA replaces the flat, full-prefix token scan with a two-stage hierarchical search (shown in Figure~\ref{fig:method}):
\begin{enumerate}
    \item \textbf{Block-level coarse filtering.} The prefix is partitioned into contiguous blocks of size $B$. A pooled representative vector is computed for each block via mean pooling over its constituent indexing keys. The query scores all $\lceil L/B \rceil$ block representatives and retains only the top-$m$ blocks, immediately pruning the majority of the prefix from further consideration.
    \item \textbf{Token-level refinement.} 
   The token-level indexer then scores at most $mB$ tokens from the candidate blocks using the same scoring mechanism as the original DSA indexer, except that the candidate pool is restricted to the tokens within the selected blocks rather than the full set of $L$ tokens considered in DSA. The final top-$k$ token set is then selected from this reduced candidate pool. 
    % The original DSA token-level indexer runs \emph{only inside the surviving candidate blocks}, scoring at most $mB$ tokens instead of $L$. The final top-$k$ token set is selected from this reduced candidate pool.
\end{enumerate}

Crucially, HISA produces outputs with the same structure as the original DSA indexer: for each query, a set of $k$ token indices. As a result, the downstream Sparse MLA operator remains entirely unchanged.
HISA is therefore a \textbf{drop-in replacement} that requires no retraining or architectural changes to the attention mechanism, and leaves the original token-level KV cache layout unchanged. HISA maintains pooled indexing-key summaries only for its coarse filter.
The per-query indexing complexity drops from $\mathcal{O}(L)$ to $\mathcal{O}(L/B + mB)$, and the per-layer cost drops from $\mathcal{O}(L^2)$ to $\mathcal{O}(L^2/B + LmB)$.

Our contributions are as follows:
\begin{itemize}
    \item We identify the indexer as an emerging bottleneck in token-level sparse attention systems and formalize the problem of \textbf{search-path optimization} for sparse indexers.
    \item We propose HISA, a hierarchical block-to-token indexing strategy that is training-free, operator-compatible, and asymptotically faster than the flat indexer.
    \item We provide optimized TileLang GPU kernel implementations for both stages of HISA and demonstrate $2$--$4\times$ kernel-level speedup at 64K contexts, as well as up to $1.65\times$ end-to-end throughput and $1.70\times$ TTFT speedup in an SGLang serving stack.
    \item We empirically validate that HISA achieves performance comparable to the original DSA on the Needle-in-a-Haystack and LongBench benchmarks.
\end{itemize}

% ===========================================================================
% 2  RELATED WORK
% ===========================================================================
\section{Related Work}
\label{sec:related}

\paragraph{Block sparse attention.}
Block sparse attention partitions sequences into fixed-size blocks and restricts computation to selected blocks, mapping naturally to GPU tiled matrix multiplications. This design is hardware-friendly, but all tokens within a block must be retained or discarded together.
Among training-free methods, MInference~\citep{jiang2024minference} profiles each head offline and assigns one of several sparse patterns at inference time; FlexPrefill~\citep{lai2025flexprefill} estimates block scores online and selects blocks by a cumulative-attention threshold; XAttention~\citep{xu2025xattention} uses antidiagonal sums as an $\mathcal{O}(B)$ proxy for block importance; and SpargeAttention~\citep{zhang2025spargeattn} applies a two-stage online filter to skip low-importance regions during matrix multiplication and softmax.
Among trainable methods, MoBA~\citep{lu2025moba} uses mixture-of-experts-style routing over blocks, while NSA~\citep{yuan2025native} combines compression, selection, and sliding-window branches to cover different dependency scales.
Their common limitation is block granularity: they cannot capture token-level importance differences within a selected block. HISA also introduces a block-level stage, but only as a fast pre-filter before token refinement; its final sparse pattern remains fine-grained and token-wise, as in DSA.

\paragraph{Token sparse attention.}
Token-level methods offer finer selection but face the challenge of efficient importance estimation.
SnapKV~\citep{li2024snapkv} uses an observation window at the end of the prompt to select important KV positions for subsequent decoding, but ignores layer- and query-specific variation.
KV cache eviction methods---such as H$_2$O~\citep{zhang2023h2o}, which combines cumulative attention with recency, and TOVA~\citep{oren2024tova}, which evicts the lowest-scoring cached token under the latest query---maintain a fixed-size cache but irrecoverably lose evicted tokens.
LazyLLM~\citep{fu2024lazyllm} progressively prunes tokens across layers during prefill, so early pruning mistakes cannot be corrected later in the same forward pass.
DSA~\citep{deepseekv32} instead scores every prefix token with a lightweight indexer and selects top-$k$ tokens per query, achieving fine-grained sparsity at the cost of $\mathcal{O}(L^2)$ per-layer indexing overhead. IndexCache~\citep{bai2026indexcache} reduces this cost by reusing indices across nearby layers, although its benefit depends on cross-layer similarity in sparse patterns.

\paragraph{Hierarchical sparse attention.}
Hierarchical attention dates back to \citet{yang2016hierarchical}, who introduced a two-tier word-and-sentence network for document classification.
Among recent sparse methods, NSA~\citep{yuan2025native} and InfLLM-V2~\citep{zhao2025infllmv2} can both be viewed as two-level designs: they score block-level summaries globally and activate finer-grained sparse attention only within selected blocks.
Twilight~\citep{lin2025twilight} uses quantized keys for coarse scoring and then applies hierarchical top-$p$ pruning, while Double-P~\citep{ni2026doublep} clusters the KV cache, scores cluster centroids, refines computation within selected clusters, and approximates low-score clusters with their centroids.
HISA follows the same coarse-to-fine spirit but with a different goal: it combines a hardware-friendly block-level indexer with a fine-grained token-level indexer to accelerate DSA, achieving both high efficiency and strong selection quality on DeepSeek-V3.2 and GLM-5.

\section{Preliminary}
\label{sec:preliminary:background}

We briefly review DeepSeek Sparse Attention (DSA) as used in DeepSeek-V3.2~\citep{deepseekv32}. DSA consists of two components: a \textbf{token-wise Indexer} and \textbf{Sparse MLA}.

\paragraph{Indexer in DSA.}
Let $L$ denote the causal prefix length for a query position $t$. The indexer maintains lightweight indexing keys $\mathbf{k}_s^I$, indexing queries $\mathbf{q}_{t,j}^I$ for $H^I$ indexing heads, and per-head gating weights $w_{t,j}^I$. The relevance score between query $t$ and key $s$ is defined as
\begin{equation}
I_{t,s} = \sum_{j=1}^{H^I} w_{t,j}^I \cdot \mathrm{ReLU}\!\left(\mathbf{q}_{t,j}^I \cdot \mathbf{k}_s^I\right).
\label{eq:dsa_score}
\end{equation}
The indexer then selects the top-$k$ token indices,
\begin{equation}
\mathcal{T}_t = \mathrm{TopK}(I_{t,:},\, k),
\label{eq:dsa_topk}
\end{equation}
which are passed to the downstream Sparse MLA operator. Since the scoring cost for each query is $\mathcal{O}(L)$ over the full prefix, the total cost across all queries in a layer is $\mathcal{O}(L^2)$.

\paragraph{Sparse MLA in DSA.}
Following the DeepSeek-V3.2 design, Sparse MLA adopts the MQA mode of MLA, in which each token stores a single latent key--value entry shared across all query heads for efficiency. Let $\mathbf{c}_s$ denote the latent MLA entry associated with token $s$. Given the selected token set $\mathcal{T}_t$, Sparse MLA computes attention for query token $t$ only over the selected latent entries, rather than over the full prefix:
\begin{equation}
\mathbf{u}_t = \mathrm{Attn}\!\left(\mathbf{h}_t,\, \left\{\mathbf{c}_s \mid s \in \mathcal{T}_t\right\}\right).
\label{eq:sparse_mla}
\end{equation}
As a result, the main attention cost is reduced from dense $\mathcal{O}(L^2)$ to sparse $\mathcal{O}(Lk)$. For our purposes, the key observation is that the interface between the two components is precisely the selected token set $\mathcal{T}_t$: HISA replaces only the indexer search path, while leaving the downstream Sparse MLA operator unchanged.

% 3  METHOD
% ===========================================================================
\section{Method}
\label{sec:method}

% ===========================================================================

\subsection{HISA: Hierarchical Indexed Sparse Attention}
\label{sec:method:hisa}

As shown in Figure \ref{fig:method}, HISA replaces the flat prefix scan with a two-stage coarse-to-fine search. The final output remains an identical per-query token set $\mathcal{T}_t^{\mathrm{H}}$ of size $k$, consumed by the original Sparse MLA operator.

\paragraph{Block partitioning and pooled keys.}
The prefix of length $L$ is partitioned into $M = \lceil L / B \rceil$ contiguous, causally valid blocks $\mathcal{B}_1, \mathcal{B}_2, \ldots, \mathcal{B}_M$, where $B$ is the block size. For each block, a representative key is constructed via mean pooling over its indexing keys:
\begin{equation}
\tilde{\mathbf{k}}_b^I = \mathrm{Pool}\!\left(\left\{\mathbf{k}_s^I \mid s \in \mathcal{B}_b\right\}\right).
\label{eq:pool}
\end{equation}

These representative keys serve exclusively as coarse-grained proxies for block-level scoring and leave both the token-level indexing keys consumed by the second stage and the KV states consumed by Sparse MLA unchanged, thereby making HISA a plug-and-play replacement. In practice, these representative keys can be incrementally maintained alongside the KV cache with negligible overhead.

\paragraph{Stage 1: Block-level coarse filtering.}
For query position $t$, HISA reuses the same indexing query representations $\mathbf{q}_{t,j}^I$ and gating weights $w_{t,j}^I$ as DSA, but scores the \textbf{pooled representative keys} instead of individual token keys:
\begin{equation}
J_{t,b} = \sum_{j=1}^{H^I} w_{t,j}^I \cdot \mathrm{ReLU}\!\left(\mathbf{q}_{t,j}^I \cdot \tilde{\mathbf{k}}_b^I\right).
\label{eq:block_score}
\end{equation}
The top-$m$ blocks sel selected:
\begin{equation}
\mathcal{C}_t = \mathrm{TopK}(J_{t,:},\, m),
\label{eq:block_topk}
\end{equation}
and the candidate token set is the union of all tokens in the selected blocks:
\begin{equation}
\Omega_t = \bigcup_{b \in \mathcal{C}_t} \mathcal{B}_b.
\label{eq:candidate}
\end{equation}
All block selections strictly respect the causal mask: only blocks that precede the query position $t$, together with the block containing position $t$, are considered eligible. Following MoBA~\citep{lu2025moba}, the first and the last blocks are \textbf{always} included in $\mathcal{C}_t$, as they contain the attention sink and local contexts. This forced inclusion also simplifies boundary handling during batched prefill with packed sequences of varying lengths, where a single block may straddle the boundary between two sequences.

\paragraph{Stage 2: Token-level refinement.}
Within the selected candidate set $\Omega_t$, the token-level indexer computes scores using the same scoring mechanism as in the original DSA (Eq.~\ref{eq:dsa_score}): 
% the original DSA token-level scoring (Eq.~\ref{eq:dsa_score}) is applied:
\begin{equation}
I_{t,s} = \sum_{j=1}^{H^I} w_{t,j}^I \cdot \mathrm{ReLU}\!\left(\mathbf{q}_{t,j}^I \cdot \mathbf{k}_s^I\right), \quad s \in \Omega_t.
\label{eq:token_refine}
\end{equation}
Then the top-$k$ tokens are selected as final tokens:
\begin{equation}
\mathcal{T}_t = \mathrm{TopK}\!\left(\left\{I_{t,s} \mid s \in \Omega_t\right\},\, k\right).
\label{eq:final_topk}
\end{equation}
% The Sparse MLA operator executes identically to the original DSA, with $\mathcal{T}_t^{\mathrm{H}}$ replacing $\mathcal{T}_t$. The feasibility constraint $mB \geq k$ must hold to ensure that the candidate pool is large enough to select $k$ tokens.

To ensure that the candidate pool is sufficiently large to select $k$ tokens, the feasibility constraint $mB \geq k$ must be satisfied. Given the selected token set $\mathcal{T}_t$, Sparse MLA is executed following the same computation as in the original DSA. 
Algorithm~\ref{alg:hisa} provides the complete pseudocode for the HISA indexer.

\paragraph{Boundary behavior.}
Three regimes arise depending on the relationship between the effective prefix length $t$, the candidate capacity $mB$, and the budget $k$:
\begin{itemize}
    \item When $t \leq k$, all prefix tokens are selected and HISA is equivalent to dense attention.
    \item When $k < t \leq mB$, the coarse filter selects all blocks (since $m \geq M$), and Stage~2 reduces the set to $k$ tokens. HISA is equivalent to the original DSA indexer.
    \item When $t > mB$, the coarse filter performs non-trivial block pruning, activating HISA's hierarchical advantage, which becomes increasingly pronounced as the sequence length grows.
\end{itemize}
The third regime is precisely the long-context setting where HISA provides its efficiency gains.

\subsection{Complexity Analysis}
\label{sec:method:complexity}

Assuming that the pooled representative keys are maintained incrementally, the per-query indexing cost of HISA consists of scoring $\lceil L/B \rceil$ block representatives (Stage~1) and scoring at most $mB$ candidate tokens (Stage~2):
\begin{equation}
\mathcal{O}\!\left(\frac{L}{B} + mB\right).
\label{eq:per_query}
\end{equation}
Summing over all $L$ queries within a layer yields:
\begin{equation}
\mathcal{O}\!\left(\frac{L^2}{B} + LmB\right),
\label{eq:per_layer}
\end{equation}
compared to $\mathcal{O}(L^2)$ for the original DSA indexer. The design introduces a clear trade-off: larger $B$ reduces the cost of the coarse-filtering stage but makes each block a coarser proxy; smaller $m$ improves efficiency but increases the risk of missing relevant blocks. When $m \ll M$ and $B \ll L$---the regime of ultra-long contexts with a selective coarse filter---the reduction is substantial. Conversely, as $m$ approaches $M$, HISA degrades gracefully toward the DSA baseline.

As modern LLMs increasingly adopt context windows of 128K or even 1M tokens to support advanced agent capabilities and native multimodal reasoning, HISA's asymptotic advantage translates directly into practical speedups.
% In today's LLM landscape, where context windows of 128K or even 1M tokens are pursued for improved agent capabilities and native multimodal reasoning, HISA's asymptotic advantage translates directly into practical speedup.

% ===========================================================================
% 4  EXPERIMENTS
% ===========================================================================

\section{Experiments}
\label{sec:experiments}
We evaluate HISA along six main axes: (1) kernel-level latency, (2) end-to-end serving performance, (3) retrieval accuracy on Needle-in-a-Haystack, (4) downstream task performance on LongBench, (5) phase sensitivity, and (6) token-selection consistency. Appendix~\ref{app:additional-results} additionally reports hyperparameter sensitivity and a failure-case analysis. Except for the native-block-indexer interventions in Section~\ref{sec:hisa-phase-sensitivity}, we compare three indexing strategies:

\begin{itemize}
    \item \textbf{DSA (original)}: the full-prefix token-level indexer as described in Section~\ref{sec:preliminary:background}.
    \item \textbf{Block-Sparse}: a block-level-only baseline that selects top-$m$ blocks and attends to all tokens within those blocks (i.e., Stage~1 only, without token-level refinement).
    \item \textbf{HISA}: the hierarchical block-to-token indexer proposed in this work.
\end{itemize}
In these comparisons, both HISA and Block-Sparse are \emph{training-free}: they are applied at inference time by replacing the indexer module, with no fine-tuning or architectural modification.

\subsection{Kernel-Level Speedup}
\label{sec:exp:kernel}

\begin{figure}[t]
\centering
\begin{subfigure}[t]{0.475\textwidth}
    \centering
    \includegraphics[width=\linewidth]{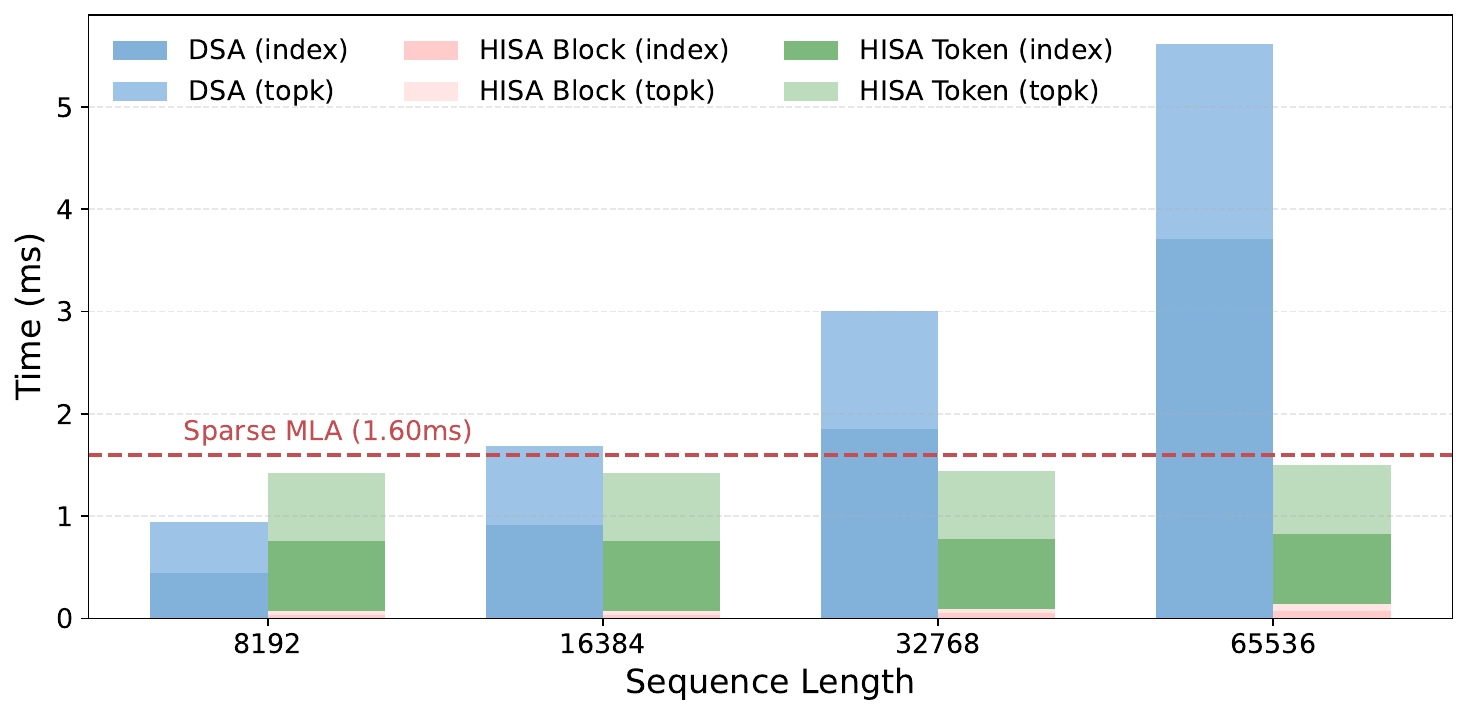}
    \caption{Budget = 8192}
    \label{fig:speed_fix_budget}
\end{subfigure}
\hfill
\begin{subfigure}[t]{0.475\textwidth}
    \centering
    \includegraphics[width=\linewidth]{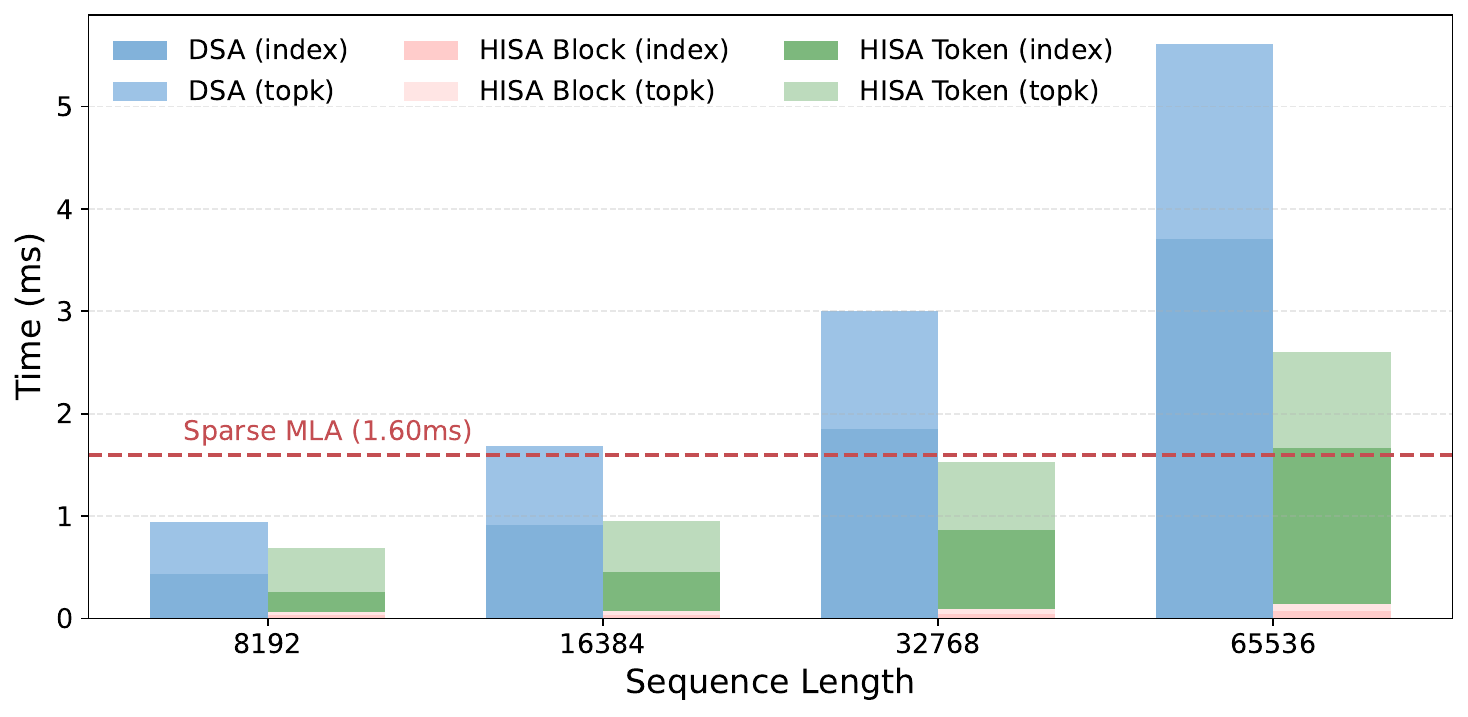}
    \caption{Compression Ratio = 4:1}
    \label{fig:speed_dynamic_budget}
\end{subfigure}
\caption{Kernel-level latency comparison of the indexer between the original DSA (flat token scan) and HISA (hierarchical block-to-token indexing). In the left panel, the block size is fixed to $B=128$ and the maximum number of selected blocks is top-$m=64$. In the right panel, $B=128$ and the number of selected blocks is adjusted for each sequence length to maintain a fixed compression ratio of $M\!:\!m = 4\!:\!1$. The red dashed horizontal line marks the latency of the (TileLang) Sparse MLA operator, which stays roughly constant across context lengths, for reference. These measurements isolate indexer time; end-to-end results, including Sparse MLA, KV-cache management, scheduling, batching, and decoding, are reported in Section~\ref{sec:exp:e2e}.}
\label{fig:speed}
\end{figure}

Figure~\ref{fig:speed} compares the indexer kernel latency of the original DSA and HISA across context lengths from 8K to 64K tokens. Both implementations use TileLang~\citep{tilelang} kernels, with DSA following the official reference implementation.\footnote{\url{https://github.com/tile-ai/tilelang/tree/main/examples/deepseek_v32}} 
The HISA kernel is decomposed into two stages: block-level filtering and token-level refinement within the selected candidate blocks. The configuration is as follows: query lens $=1024$, final top-$k=2048$ tokens, block size $B=128$, and two choices for the maximum number of selected blocks. All comparisons are conducted on an NVIDIA A100 GPU. These results are measured at the \emph{indexer kernel} level and do not directly reflect end-to-end serving throughput, which also depends on the Sparse MLA operator, KV cache management, and other system components.

With 2048 selected tokens, the Sparse MLA operator consistently costs about 1.6 ms, while the indexer reaches 5.6 ms at 64K context length. This suggests that the main performance bottleneck in DSA lies in the indexer rather than in Sparse MLA itself. Accordingly, this section isolates indexer latency to characterize this bottleneck in a controlled setting; Section~\ref{sec:exp:e2e} reports the corresponding end-to-end serving results.
At 64K context length, HISA delivers an approximately $2.16\times$ speedup with a 4:1 first-stage compression ratio (corresponding to a 16K candidate budget), and up to $3.75\times$ speedup under a fixed 8K budget. Although HISA adds a block-level filtering stage, this stage operates only on pooled block summaries of size $\lceil L/B \rceil$, which is far smaller than the full token sequence. Moreover, under a fixed 8K budget, the second-stage cost remains nearly constant because both the input and output lengths are fixed, making the computation graph easier to optimize and further improving inference speed.

\subsection{End-to-End Serving Performance}
\label{sec:exp:e2e}

We evaluate HISA in the full SGLang serving stack using DeepSeek-V3.2, against the official DeepSeek-V3.2 DSA implementation in SGLang as baseline; Appendix~\ref{app:e2e-settings} details the hardware and workload configuration. The baseline uses DeepSeek's hand-tuned CUDA kernels, including \texttt{deep\_gemm} for MQA indexer scoring, while HISA retains the optimized downstream \texttt{flashmla\_sparse} path and replaces only the indexer; its newly introduced block pooling and sparse-MQA operators are currently implemented in TileLang or Triton and have not received the same degree of low-level tuning as the baseline.

We measure end-to-end serving on SGLang rather than vLLM, which we use for the quality evaluations in Sections~\ref{sec:exp:niah} and \ref{sec:exp:longbench}, because SGLang's serving architecture more readily accommodates HISA's Stage-1 block-pooled key cache alongside the original token-level KV cache. Appendix~\ref{app:sglang-settings} gives the full reasoning, and Appendix~\ref{app:sglang-vllm} reports a LongBench cross-check confirming that HISA's quality conclusions transfer between the two stacks.

\begin{table}[t]
\centering
\caption{End-to-end serving performance on DeepSeek-V3.2 in SGLang ($8\times$H200, TP=8, 128K input and 512 output tokens per request). HISA achieves $1.65\times$ throughput and $1.70\times$ TTFT speedup despite its TileLang/Triton operators being less mature than the official DSA baseline's hand-tuned CUDA kernels; the measured speedups therefore understate HISA's potential at matched kernel maturity.}
\label{tab:e2e}
\small
\begin{tabular}{lrrrr}
\toprule
\textbf{Indexer} & \textbf{Throughput} & \textbf{E2E latency} & \textbf{TTFT} & \textbf{TPOT} \\
& \textbf{(tok/s)} & \textbf{(ms)} & \textbf{(ms)} & \textbf{(ms)} \\
\midrule
DSA (official) & 5868.43 & 143764.47 & 106449.81 & 73.02 \\
HISA ($B{=}128, m{=}64$) & 9712.21 & 86718.06 & 62722.91 & 46.96 \\
\midrule
Speedup & $1.65\times$ & $1.66\times$ & $1.70\times$ & $1.55\times$ \\
\bottomrule
\end{tabular}
\end{table}

Table~\ref{tab:e2e} shows that HISA improves all system-level metrics: throughput, end-to-end latency, time to first token (TTFT), and time per output token (TPOT). Unlike the kernel microbenchmark, this measurement includes Sparse MLA, KV-cache management, scheduling, batching, and autoregressive decoding. Since the HISA-specific operators have not yet received CUDA-level tuning comparable to the official DSA kernels, these end-to-end gains are a conservative estimate of HISA's benefit at matched kernel maturity. The results nevertheless demonstrate that reducing index-selection cost translates into end-to-end serving gains on top of an optimized sparse-attention backend.

The gains persist across request concurrencies; detailed scaling results are provided in Appendix~\ref{app:e2e-concurrency}.

\subsection{Needle-in-a-Haystack}
\label{sec:exp:niah}

\begin{figure}[t]
\centering
\begin{subfigure}[t]{0.29\textwidth}
    \centering
    \includegraphics[width=\linewidth]{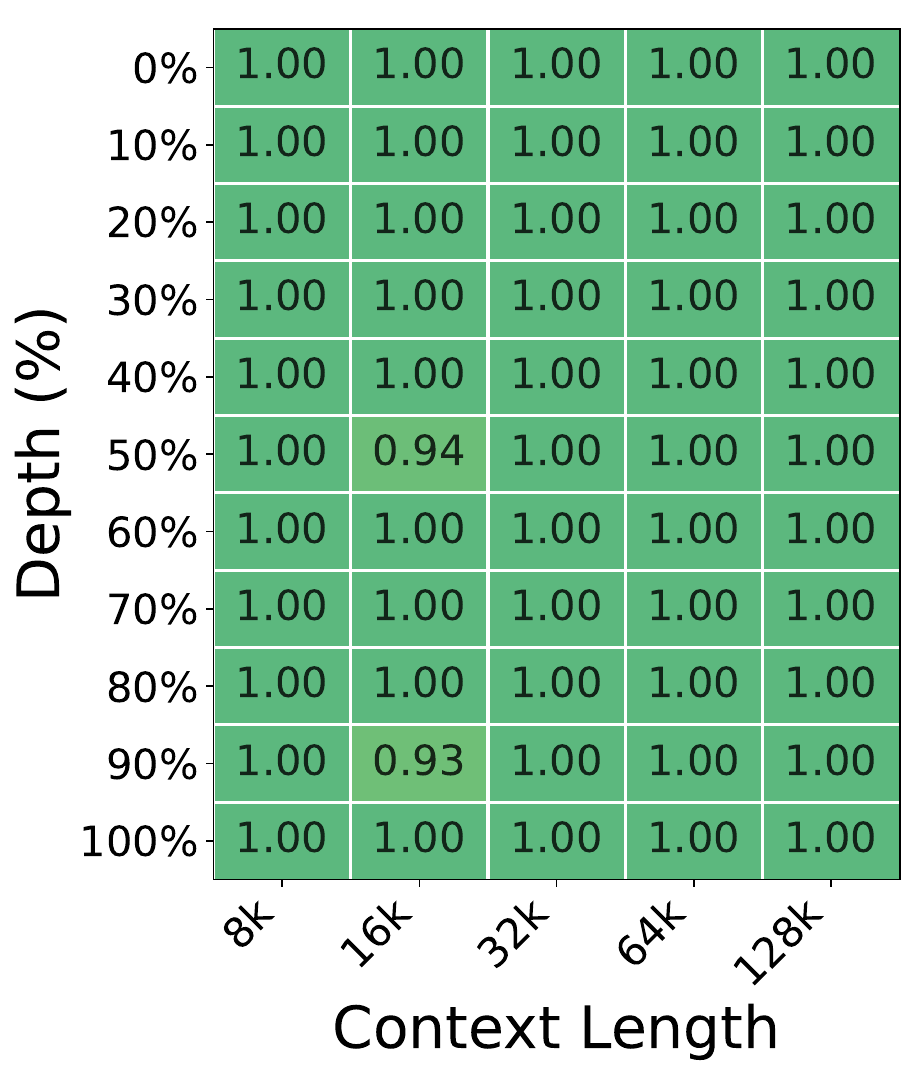}
    \caption{DSA (original)}
    \label{fig:niah_dsa}
\end{subfigure}
\hfill
\begin{subfigure}[t]{0.29\textwidth}
    \centering
    \includegraphics[width=\linewidth]{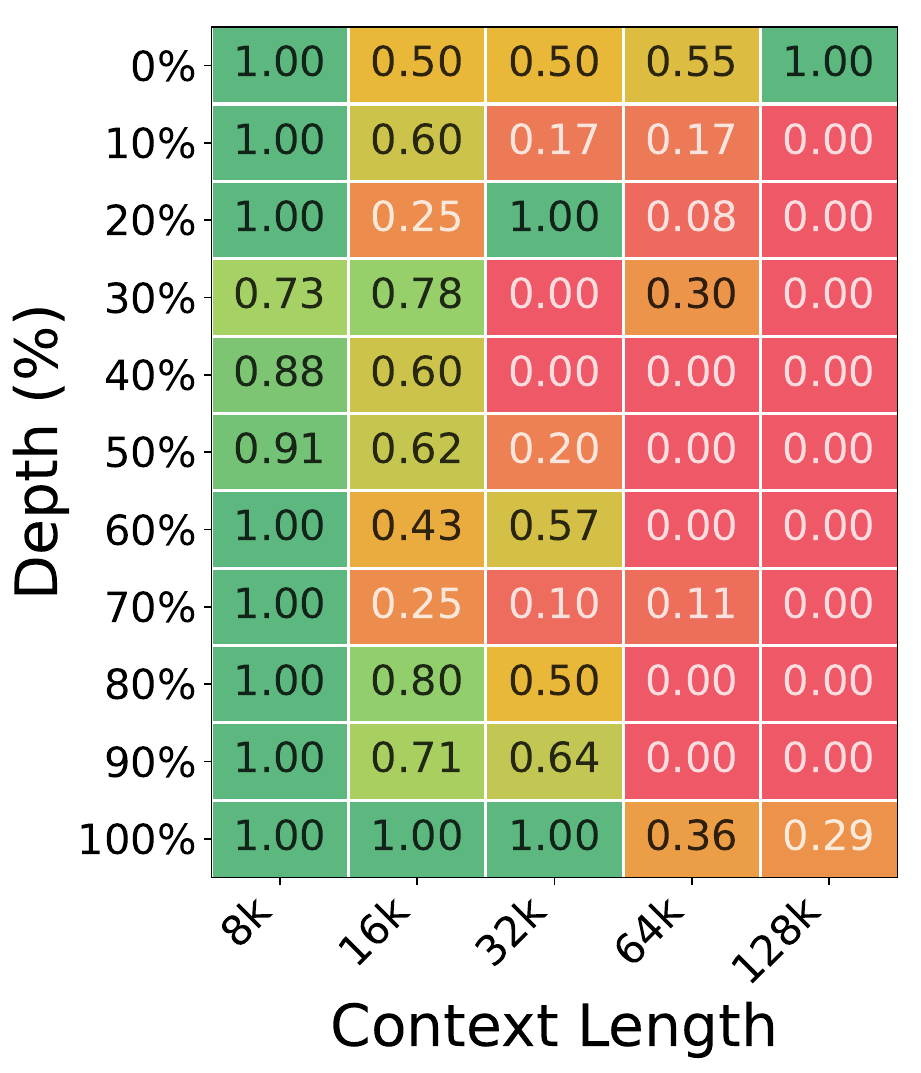}
    \caption{Block-Sparse}
    \label{fig:niah_block}

\end{subfigure}
\hfill
\begin{subfigure}[t]{0.36\textwidth}
    \centering
        \includegraphics[width=\linewidth]{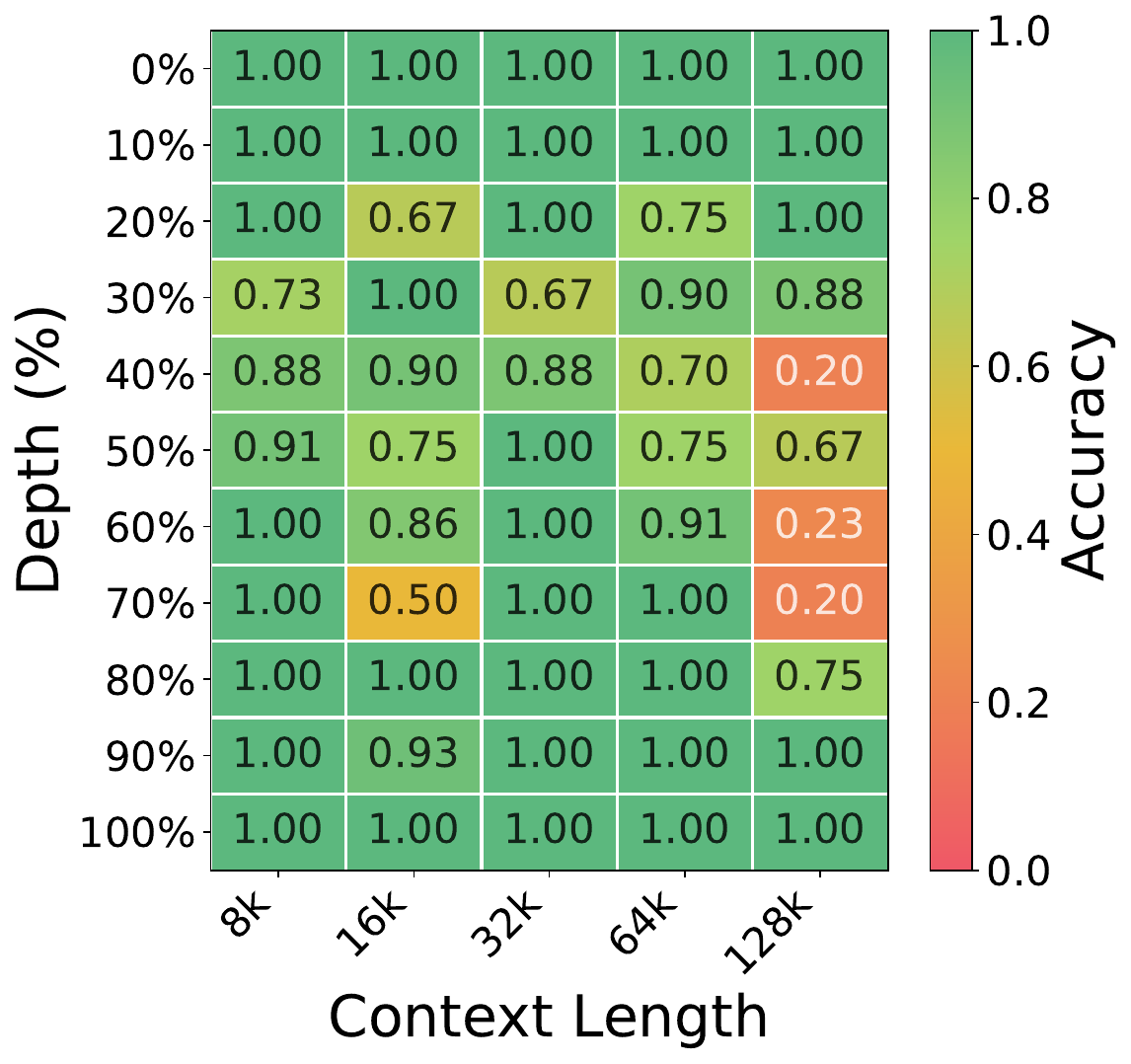}
    \caption{HISA (ours)}
    \label{fig:niah_hisa}
\end{subfigure}
\caption{Needle-in-a-Haystack retrieval accuracy heatmaps for DeepSeek-V3.2 under three indexing strategies. The $x$-axis denotes the context length (8K--128K), and the $y$-axis denotes the needle depth (0\%--100\%). Shades closer to green indicate higher retrieval accuracy.}
\label{fig:niah}
\end{figure}

The Needle-in-a-Haystack (NIAH) test~\citep{kamradt2023needle} evaluates a model's ability to retrieve a specific fact (the "needle") embedded at a controlled position within a long distractor context (the "haystack"). We evaluate DeepSeek-V3.2 with its original DSA indexer replaced by HISA (4:1 ratio) and block indexer, without any additional training, over context lengths ranging from 8K to 648K tokens and needle insertion depths ranging from 0\% (beginning) to 100\% (end).

Figure~\ref{fig:niah} presents the retrieval accuracy heatmaps. The original DSA achieves near-perfect retrieval across all context lengths and needle positions (Figure~\ref{fig:niah_dsa}). HISA closely matches this performance (Figure~\ref{fig:niah_hisa}), with only marginal degradation at extreme lengths and depths, suggesting that HISA rarely discards blocks containing the target information. In contrast, the Block-Sparse baseline (Figure~\ref{fig:niah_block}) exhibits noticeable accuracy degradation, particularly when the needle is located in the middle of the context where block-level selection is least reliable. This result underscores the value of hierarchical selection. Block-sparse methods often waste budget on unimportant tokens within selected blocks while overlooking truly critical tokens. HISA, in contrast, refines the selection at the token level after block retrieval, allowing it to preserve important tokens more accurately and achieve efficient token-wise sparsity.

\subsection{LongBench Evaluation}
\label{sec:exp:longbench}

LongBench~\citep{bai2024longbench} is a comprehensive benchmark for long-context understanding, covering single-document QA, multi-document QA, summarization, few-shot learning, and synthetic retrieval tasks. We evaluate DeepSeek-V3.2~\citep{deepseekv32} and GLM-5~\citep{glm5} under three configurations: the original DSA indexer, HISA, and Block-Sparse Attention. 
For a fair comparison, all three configurations ultimately retain 2048 tokens for computation. Specifically, Block-Sparse Attention directly selects 16 blocks of size 128 (i.e., $128 \times 16 = 2048$ tokens). HISA first selects 64 blocks of size 128 (i.e., $128 \times 64 = 8192$ tokens), and then further refines them through token-level selection to 2048 tokens.

\begin{table}[t]
\centering
\caption{LongBench results for DeepSeek-V3.2 and GLM-5 under different indexing strategies. All sparse methods are applied at inference time without additional training. Scores are averaged across sub-tasks within each category. Task abbreviations: \textbf{SQA}~=~Single-Document QA, \textbf{MQA}~=~Multi-Document QA, \textbf{Sum}~=~Summarization, \textbf{FS}~=~Few-shot Learning, \textbf{Syn}~=~Synthetic Retrieval, \textbf{Code}~=~Code Completion.}
\label{tab:longbench}
\vspace{0.5em}
\begin{tabular}{ll cccccc|c}
\toprule
\textbf{Model} & \textbf{Indexer} & \textbf{SQA} & \textbf{MQA} & \textbf{Sum} & \textbf{FS} & \textbf{Syn}& \textbf{Code} & \textbf{Avg.} \\
\midrule
\multirow{3}{*}{DeepSeek-V3.2}
& DSA   & \textbf{50.89} & \textbf{52.66} & \underline{22.11} & \textbf{62.24} & \underline{69.83} & 48.56 & \textbf{51.05} \\
& Block & 48.36 & 49.76 & 21.90 & 59.45 & 68.67 & \textbf{49.09} & 49.54 \\
& HISA  & \underline{49.17} & \underline{51.96} & \textbf{22.13} & \underline{61.62} & \textbf{70.83} & \underline{48.99} & \underline{50.78} \\
\midrule
\multirow{3}{*}{GLM-5}
& DSA   & \underline{41.23} & \textbf{27.89} & \textbf{18.39} & \underline{63.20} & \underline{68.84} & \underline{56.53} & \underline{46.01} \\
& Block & 38.35 & 24.29 & 16.95 & 60.64 & 60.49 & 55.29 & 42.67 \\
& HISA  & \textbf{42.45} & \underline{27.62} & \underline{17.90} & \textbf{63.78} & \textbf{69.35} & \textbf{56.79} & \textbf{46.32} \\
\bottomrule
\end{tabular}%
\end{table}

Table~\ref{tab:longbench} summarizes the results. Across both models and all task categories, HISA achieves performance very close to that of the original DSA. Notably, HISA consistently surpasses DSA on the Synthetic tasks, and on GLM-5 it even attains a higher average score.
By contrast, the Block-Sparse baseline, which does not include token-level refinement, exhibits a substantially larger performance gap. This is particularly apparent on the Synthetic tasks for GLM-5, where its score declines by 8.35 points.

\subsection{Phase Sensitivity under Block-to-Token Indexing}
\label{sec:hisa-phase-sensitivity}

\begin{figure}[t]
  \centering
  \includegraphics[width=\linewidth]{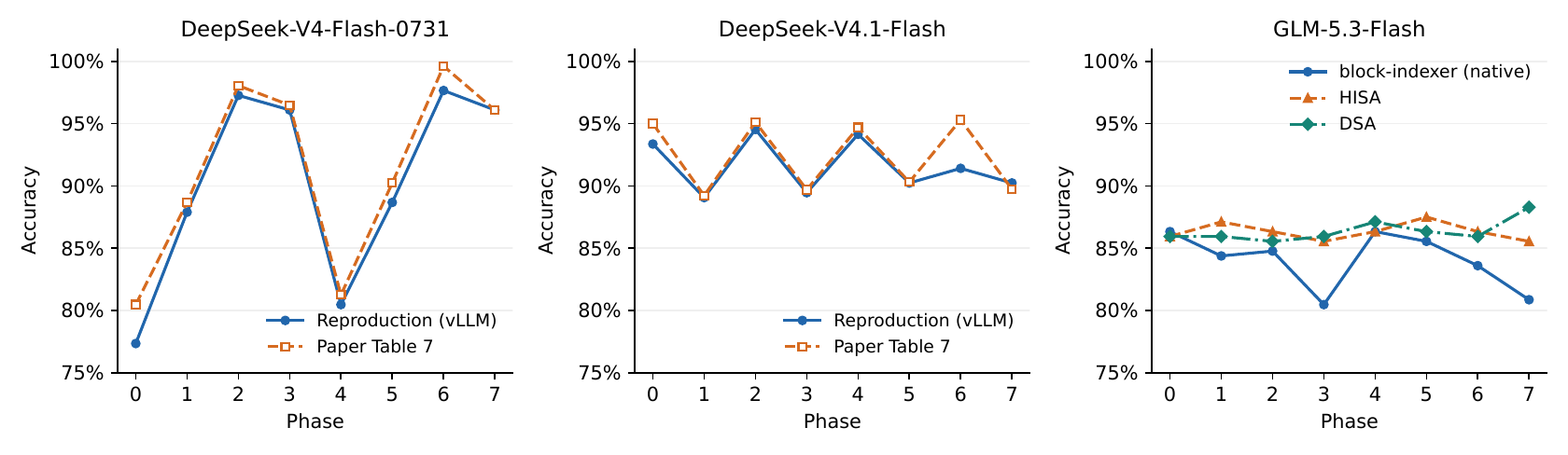}
  \caption{Retrieval accuracy by source-position phase modulo eight.
  Left and middle: our vLLM reproductions and the published Table~7 values of \citet{zhu2026periodicweakspots}.
  Right: GLM-5.3-Flash with its native block indexer and training-free HISA/DSA interventions.
  DeepSeek uses the original answer-prefix metric; GLM uses the quote-tolerant metric.
  Curves show point estimates; the shared, truncated vertical axis does not imply that absolute scores are comparable across tokenizers and scoring rules.}
  \label{fig:hisa-phase-sensitivity}
\end{figure}

Chunk-based indexing and compression can introduce position-dependent retrieval errors.
Using the released code of \citet{zhu2026periodicweakspots}, we reproduce the four-phase and two-phase accuracy patterns of DeepSeek-V4-Flash-0731 and DeepSeek-V4.1-Flash, respectively, consistent with their CSA-4 and CSA-2 strides (Figure~\ref{fig:hisa-phase-sensitivity}, left and middle).
Because DeepSeek's compressed sparse attention couples indexer compression with compression of the downstream attention KV representations, these observations alone do not isolate the contribution of selection granularity.

\paragraph{Controlled indexer interventions.}
We therefore extend the retrieval experiment to GLM-5.3-Flash.
In its sparse-attention layers, the native indexer pools four tokens into each indexing block, while downstream attention operates on the selected original token-level KV representations.
We keep the pretrained weights and downstream attention unchanged and compare three selection rules across all 11 sparse layers:
(i) \textbf{block-indexer (native)}, which selects 512 four-token blocks;
(ii) \textbf{DSA}, which scores individual tokens across the causal history and selects the top 2,048;
and (iii) \textbf{HISA}, which first retrieves 2,048 four-token blocks (8,192 candidate tokens), then uses the same token scorer as DSA to select the final 2,048 tokens.
HISA thus uses the native four-token blocks for this diagnostic, rather than the 128-token blocks of our preceding LongBench experiment.
Both interventions are training-free and reuse the original indexer projections and head weights.
Native handling of an unfinished block can add up to three tokens; these tail candidates compete within the strict final 2,048-token budget for HISA (Appendix~\ref{app:hisa-phase-details}).

\paragraph{Protocol and scoring.}
Each configuration is evaluated on 2,048 prompts covering eight source-position phases, with the same 256 key--value bindings evaluated at every phase.
All GLM variants receive identical prompts, with 128,000 tokens before any tokenizer-added special tokens, and use greedy decoding for at most eight output tokens without a chat template.
We retain the authors' answer-prefix metric for DeepSeek.
For GLM, we report quote-tolerant answer-prefix accuracy, which additionally removes leading quotation marks; this diagnostic was introduced after observing output-format errors.
The original GLM metric and the exact normalization rule are reported in Appendix~\ref{app:hisa-phase-details}.
The native reference is a control rerun with the same runtime settings as the interventions.

\paragraph{Results and interpretation.}
The native GLM indexer exhibits recurring troughs at phases 3 and 7 despite retaining token-level KV representations in downstream sparse attention.
Its overall accuracy is 84.03\%, with a 5.86-percentage-point maximum gap across the eight phases.
HISA and DSA attain 86.33\% and 86.38\%, while reducing the observed gaps to 1.95 and 2.73 points, respectively.
Thus, the flatter curves do not arise from a collapse in quote-tolerant retrieval accuracy.
The accuracy gain is metric-dependent: under the original rule, HISA scores 73.19\% versus 76.81\% for the native control.
These observations support token-level refinement as a way to mitigate alignment-dependent retrieval errors: block-level coarse filtering need not impose block-level final selection.
They also suggest that compression of downstream attention KV representations is not necessary for such errors to occur, without identifying the unique cause of DeepSeek's behavior.
We describe the effect as mitigation rather than elimination: paired-bootstrap 95\% intervals for the reduction in the eight-phase gap include zero (HISA: $[-1.17,7.81]$ points; DSA: $[-1.56,7.81]$ points).
This experiment evaluates retrieval behavior, not the latency of an optimized HISA implementation.

\subsection{Visualization of Attention Scores}
\label{sec:exp:att-visual}
To analyze the structural properties of attention in long-context generation, we conduct a visualization study on a representative sample from the code task of LongBench. We generate the first output token using DeepSeek-V3.2 and extract the full attention distributions at each layer. We visualize the attention weights over all context tokens as a 2D heatmap, where the x-axis denotes token positions and the y-axis denotes layer indices.

\begin{figure}[t]
\centering
\includegraphics[width=1.0\linewidth]{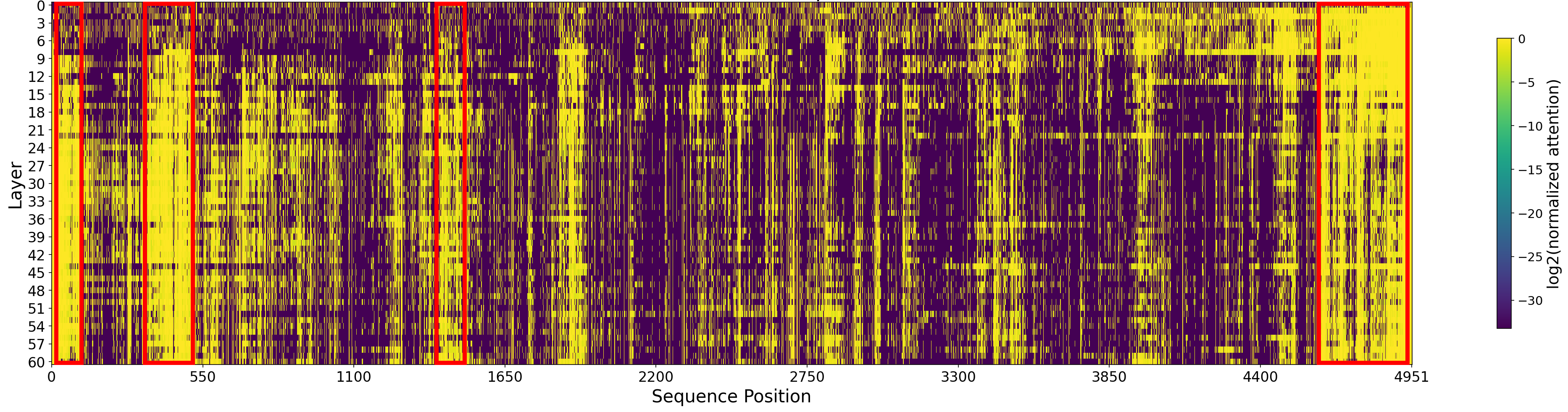}
\caption{Visualization of Attention Distribution.}
\label{fig:att-visual}
\end{figure}

The visualization reveals a pattern: tokens with high attention weights tend to form contiguous spans rather than appearing as isolated points in a considerable number of tasks. These high-density regions often correspond to semantically coherent segments (e.g., code blocks, mathematical formulas and derivations) and persist across multiple layers. Outside these spans, attention scores are negligible.
This observation suggests that attention mass may be naturally concentrated in block-wise regions. Therefore, block-level sparsification can retain most of the informative attention distribution while avoiding the fine-grained selection overhead of token-wise top-k methods. The results provide empirical support for the two-stage hierarchical structure of HISA.

\subsection{Selection Consistency: IoU Analysis}
\label{sec:exp:iou}

To quantify how closely HISA recovers DSA's token selections, we measure the Intersection-over-Union (IoU) between their selected token sets under the deployment configuration $B=128, m=64$:
\begin{equation}
\mathrm{IoU}\!\left(\mathcal{T}_t^{\mathrm{H}},\, \mathcal{T}_t\right) = \frac{\left|\mathcal{T}_t^{\mathrm{H}} \cap \mathcal{T}_t\right|}{\left|\mathcal{T}_t^{\mathrm{H}} \cup \mathcal{T}_t\right|}.
\label{eq:iou}
\end{equation}

\begin{table}[t]
\centering
\caption{Token-set IoU (Eq.~\ref{eq:iou}) B $B{=}128, m{=}64$, binned in two complementary ways. (a) Averaged over 10-layer bins across all 61 layers, IoU is uniformly high (92.58\%--94.94\%) with no systematic trend across depth, indicating that HISA's coarse filter agrees with DSA's token selection at every layer. (b) Averaged over 2K-token query-position bins, IoU starts at 96.91\% for the earliest queries (2--4K) and declines monotonically to 78.66\% at the longest evaluated positions (18.4--18.8K): as the prefix lengthens, more candidate blocks compete for the fixed top-$m$ budget, so Stage~1 is more likely to disagree with DSA's exhaustive token-level scoring on which blocks contain the true top-$k$ tokens.}
\label{tab:iou}
\small
\textbf{(a) Averaged over layers}\\[0.3em]
\begin{tabular*}{\linewidth}{@{\extracolsep{\fill}}lrrrrrrr}
\toprule
\textbf{Layer range} & 0--9 & 10--19 & 20--29 & 30--39 & 40--49 & 50--59 & 60 \\
\midrule
\textbf{IoU (\%)} & 94.90 & 94.94 & 93.43 & 92.58 & 92.85 & 93.21 & 93.94 \\
\bottomrule
\end{tabular*}
\vspace{0.7em}

\textbf{(b) Averaged over query token position}\\[0.3em]
\begin{tabular*}{\linewidth}{@{\extracolsep{\fill}}lrrrrr}
\toprule
\textbf{Position range} & 2--4K & 4--6K & 6--8K & 8--10K & 10--12K \\
\midrule
\textbf{IoU (\%)} & 96.91 & 94.74 & 93.83 & 91.88 & 88.09 \\
\midrule
\textbf{Position range} & 12--14K & 14--16K & 16--18K & 18--18.8K & \\
\midrule
\textbf{IoU (\%)} & 85.07 & 86.47 & 79.28 & 78.66 & \\
\bottomrule
\end{tabular*}
\end{table}

The consistently high layer-wise overlap and the graceful position-wise decay in Table~\ref{tab:iou} are consistent with HISA's LongBench accuracy staying close to DSA overall (Table~\ref{tab:longbench}) while occasionally diverging on the hardest, longest-context queries. HISA therefore preserves the same token-index output contract for Sparse MLA while approximating, rather than exactly reproducing, DSA's selected token set. Appendix~\ref{app:additional-results} further reports a failure-case analysis and a hyperparameter sensitivity study, including an ablation over substantially larger block sizes.

% ===========================================================================
% 6  CONCLUSION
% ===========================================================================
\section{Conclusion and Future Directions}
\label{sec:conclusion}

To address the emerging bottleneck caused by the $O(L^2)$ complexity of the DSA indexer, we propose HISA, a hierarchical indexing approach. Specifically, HISA first uses a hardware-friendly block indexer to efficiently filter out a large number of irrelevant tokens, and then applies token-level reranking over the remaining candidates to construct the token indices consumed by Sparse MLA. HISA delivers up to $3.75\times$ indexer-kernel speedup over DSA at 64K context.
As a plug-and-play module, HISA can directly replace the token indexer in DeepSeek-V3.2 and GLM-5. Without additional training, it closely matches DSA on LongBench and substantially outperforms the corresponding block-sparse baseline on NIAH. In a full SGLang serving stack, HISA improves throughput by $1.65\times$, TTFT by $1.70\times$, and TPOT by $1.55\times$ over the official DSA implementation.

Several avenues remain open: (1)~\emph{Reducing information loss in coarse filtering}: the current block-level stage represents each block with a single pooled vector, which can fail when a block crosses a semantic boundary and the pooled representation does not reflect the most important token. Potential mitigations include overlapping blocks, adaptive block boundaries, or replacing mean pooling with max pooling to better preserve salient outlier directions. (2)~\emph{Training-aware HISA}: while HISA currently operates as a training-free inference-time replacement, jointly training the block scoring stage may improve the coarse filter's accuracy, particularly for such boundary cases. (3)~\emph{Serving optimization}: fusing the stages and tuning the newly introduced operators for small block top-$m$ could further reduce the remaining serving overhead.

\section*{Acknowledgments}

This work was partially supported by PKU Kunpeng \& Ascend Center of Excellence and the CCF-Tencent Rhino-Bird Open Research Fund.

\newpage
% ===========================================================================
% REFERENCES
% ===========================================================================
\bibliography{colm2026_conference}
\bibliographystyle{colm2026_conference}
\newpage
% ===========================================================================
% APPENDIX
% ===========================================================================
\appendix
\section{Algorithm Pseudocode}
\label{app:algorithm}

Algorithm~\ref{alg:hisa} provides the complete pseudocode for the HISA indexer.

\begin{algorithm}[h]
\caption{HISA: Hierarchical Indexed Sparse Attention}
\label{alg:hisa}
\begin{algorithmic}[1]
\REQUIRE Query indexing representations $\{\mathbf{q}_{t,j}^I\}$, gating weights $\{w_{t,j}^I\}$, token indexing keys $\{\mathbf{k}_s^I\}_{s=1}^{L}$, block size $B$, block budget $m$, token budget $k$
\ENSURE Selected token set $\mathcal{T}_t$ of size $k$
\STATE Partition prefix into $M = \lceil L/B \rceil$ blocks $\mathcal{B}_1, \ldots, \mathcal{B}_M$
\FOR{$b = 1$ to $M$}
    \STATE $\tilde{\mathbf{k}}_b^I \leftarrow \mathrm{MeanPool}(\{\mathbf{k}_s^I \mid s \in \mathcal{B}_b\})$
\ENDFOR
\FOR{each query position $t$}
    \STATE \textit{// Stage 1: Block-level coarse filter}
    \FOR{$b = 1$ to $M$ \textbf{such that} $\mathcal{B}_b$ precedes $t$}
        \STATE $J_{t,b} \leftarrow \sum_{j} w_{t,j}^I \cdot \mathrm{ReLU}(\mathbf{q}_{t,j}^I\cdot \tilde{\mathbf{k}}_b^I)$
    \ENDFOR
    \STATE $\mathcal{C}_t \leftarrow \mathrm{TopK}(J_{t,:},\, m) \cup \{\text{first block, last block}\}$
    \STATE $\Omega_t \leftarrow \bigcup_{b \in \mathcal{C}_t} \mathcal{B}_b$
    \STATE \textit{// Stage 2: Token-level refinement}
    \FOR{$s \in \Omega_t$}
        \STATE $I_{t,s} \leftarrow \sum_{j=1}^{H^I} w_{t,j}^I \cdot \mathrm{ReLU}(\mathbf{q}_{t,j}^I \cdot \mathbf{k}_s^I)$
    \ENDFOR
    \STATE $\mathcal{T}_t \leftarrow \mathrm{TopK}(\{I_{t,s} \mid s \in \Omega_t\},\, k)$
\ENDFOR
\RETURN $\mathcal{T}_t$
\end{algorithmic}
\end{algorithm}

\section{Experimental Settings}

We detail additional experimental settings in this section. All long-context evaluations were conducted in a \textbf{zero-shot} setting.

\subsection{End-to-End Serving Settings}
\label{app:e2e-settings}

We evaluate end-to-end serving (Section~\ref{sec:exp:e2e}) on DeepSeek-V3.2 with $8\times$ NVIDIA H200 GPUs and tensor parallelism $8$. The workload consists of 10 random requests, each with 128K input tokens and 512 generated tokens; all numbers are measured after JIT warm-up.

\subsection{End-to-End Serving: Why SGLang}
\label{app:sglang-settings}

We measure end-to-end serving on SGLang rather than vLLM because SGLang exposes an explicit request-to-token-pool abstraction that cleanly accommodates HISA's heterogeneous KV layout: the original token-level KV cache alongside HISA's Stage-1 block-pooled key cache, which stores one pooled indexing key per block rather than per token. vLLM's cache classes, in contrast, enforce a uniform page size across cache classes, which conflicts with this block-pooled cache having a different granularity from the token-level cache. Since this changes the serving stack used to measure speed relative to the quality evaluations in Sections~\ref{sec:exp:niah} and \ref{sec:exp:longbench} (both on vLLM), Appendix~\ref{app:sglang-vllm} reports a LongBench cross-check confirming that HISA's quality conclusions transfer between the two stacks.

\subsection{Long-context Benchmarks}
We evaluated the long-context performance using the Needle In A Haystack (NIAH) test and the LongBench benchmark. We tested two models: \textbf{DeepSeek-V3.2} and \textbf{GLM-5}. Both models were deployed using the vLLM online serving framework with \textbf{FP8} precision.

\paragraph{NIAH Settings}
For the NIAH experiments, we utilized a customized evaluation codebase modified from the RULER\footnote{https://github.com/NVIDIA/RULER} GitHub repository. We did not apply chat templates to either model to ensure a direct assessment of their raw retrieval capabilities.

\paragraph{LongBench Settings}
We evaluated LongBench using the lm-eval\footnote{https://github.com/EleutherAI/lm-evaluation-harness} framework. The configurations for LongBench varied slightly depending on the model characteristics:
\begin{itemize}
    \item \textbf{Chat Template Usage:} 
    DeepSeek-V3.2 was evaluated with its standard chat template. 
    In contrast, GLM-5 was evaluated \textit{without} a chat template. This decision was made because using the template triggered an extended thinking process that exceeded the maximum generation length and significantly slowed down inference. Furthermore, disabling the thinking process while keeping the template resulted in inferior performance compared to not using the template at all.
    
    \item \textbf{Concurrency Settings:}
    The default number of concurrent requests (\texttt{num\_concurrent}) was set to 20. However, due to Out-Of-Memory (OOM) issues specific to GLM-5 on certain tasks, we adjusted the concurrency: \texttt{longbench\_single} was run with a concurrency of 1, and \texttt{longbench\_summary} was run with a concurrency of 2.
\end{itemize}

\paragraph{Fairness of Comparison}
We emphasize that although the specific settings (e.g., concurrency, chat template) differ across models and tasks to accommodate their unique characteristics and hardware constraints, we ensure that the settings are \textbf{strictly aligned} when comparing different methods within the same model and task combination. This guarantees a fair and rigorous comparison.

\section{Additional Experimental Results}
\label{app:additional-results}

\subsection{End-to-End Serving Concurrency}
\label{app:e2e-concurrency}

To assess whether HISA's gains persist under batched serving, Table~\ref{tab:e2e-concurrency} extends the Section~\ref{sec:exp:e2e} evaluation to request concurrencies of 1, 2, and 4. The trends are consistent with a fixed per-request TTFT saving that becomes easier to realize in TPOT as batching increases GPU occupancy.

\begin{table}[t]
\centering
\caption{End-to-end speedup of HISA ($B{=}128, m{=}64$) over the official DSA baseline in SGLang across request concurrencies $1$, $2$, and $4$, under the same DeepSeek-V3.2, $8\times$H200 (TP=8), 128K-input / 512-output workload as Table~\ref{tab:e2e}. TTFT speedup stays essentially flat at $1.75$--$1.76\times$ across all concurrencies, since HISA's reduction in first-token indexing cost is a per-request property largely independent of batching. TPOT speedup instead grows markedly with concurrency, from $1.05\times$ at concurrency~1 to $1.55\times$ at concurrency~4: at low concurrency, per-step decoding latency is dominated by fixed serving overhead and low GPU occupancy that indexer speedup alone cannot offset, whereas higher concurrency exposes more of the decode-side indexing cost, making HISA's saving more visible in TPOT. We report up to concurrency~4 because, under this 128K-context workload, the KV cache in our serving setup can hold at most four concurrent requests.}
\label{tab:e2e-concurrency}
\small
\begin{tabular}{crrrr}
\toprule
\textbf{Concurrency} & \textbf{Throughput} & \textbf{E2E latency} & \textbf{TTFT} & \textbf{TPOT} \\
\midrule
1 & $1.48\times$ & $1.48\times$ & $1.76\times$ & $1.05\times$ \\
2 & $1.58\times$ & $1.58\times$ & $1.76\times$ & $1.38\times$ \\
4 & $1.64\times$ & $1.65\times$ & $1.75\times$ & $1.55\times$ \\
\bottomrule
\end{tabular}
\end{table}

\subsection{SGLang vs.\ vLLM Quality Consistency}
\label{app:sglang-vllm}

Section~\ref{sec:exp:e2e} measures end-to-end serving speed on SGLang, while the quality evaluations in Sections~\ref{sec:exp:niah} and \ref{sec:exp:longbench} use vLLM. Since the HISA algorithm itself is unchanged across the two stacks, we re-run a subset of LongBench on SGLang to confirm that our quality conclusions carry over to the stack used for the E2E measurements.

\begin{table}[h]
\centering
\caption{LongBench results for DeepSeek-V3.2 on the SGLang stack, alongside the corresponding vLLM averages from Table~\ref{tab:longbench} for reference. The same qualitative trend holds on both stacks: DSA and HISA remain close to each other while both clearly outperform Block-Sparse.}
\label{tab:sglang-vllm}
\small
\begin{tabular}{lrrrrrrr}
\toprule
\textbf{Config} & \textbf{SQA} & \textbf{MQA} & \textbf{Sum} & \textbf{FS} & \textbf{Syn} & \textbf{Avg.} & \textbf{vLLM Ref.} \\
\midrule
DSA (official) & 49.79 & 52.68 & 22.24 & 60.29 & 66.69 & 50.34 & 51.55 \\
HISA ($B{=}128, m{=}64$) & 49.26 & 51.82 & 22.07 & 59.75 & 67.33 & 50.05 & 51.14 \\
Block-Sparse ($B{=}128, m{=}16$) & 46.99 & 50.03 & 21.46 & 55.24 & 66.14 & 47.97 & 49.63 \\
\bottomrule
\end{tabular}
\end{table}

On SGLang, DSA (50.34) and HISA (50.05) remain close while both clearly outperform Block-Sparse (47.97), matching the vLLM trend reported in Table~\ref{tab:longbench} (DSA 51.55, HISA 51.14, Block-Sparse 49.63). The two stacks are not expected to produce identical scores---they differ in scheduling, batching, and numerical details unrelated to HISA itself---but the ranking and relative gaps are consistent, supporting the use of SGLang for the E2E speed measurements in Section~\ref{sec:exp:e2e} without re-validating quality from scratch on that stack.

\subsection{Hyperparameter Sensitivity}
\label{app:hyperparams}

We investigate the sensitivity of HISA to its two key hyperparameters---block size $B$ and block-level top-$m$---by comparing three HISA configurations that share the same candidate pool size, $mB = 8192$, but different coarse-to-fine trade-offs: $(B{=}64, m{=}128)$, $(B{=}128, m{=}64)$, and $(B{=}256, m{=}32)$. We further include the original DSA as an upper bound and Block-Sparse $(B{=}128, m{=}16)$ as a lower bound. All configurations use $k{=}2048$ for the final token selection. Results are evaluated on DeepSeek-V3.2 and GLM-5 across five LongBench task categories.

\begin{figure}[h]
\centering
\begin{subfigure}[t]{\textwidth}
    \centering
    \includegraphics[width=\linewidth]{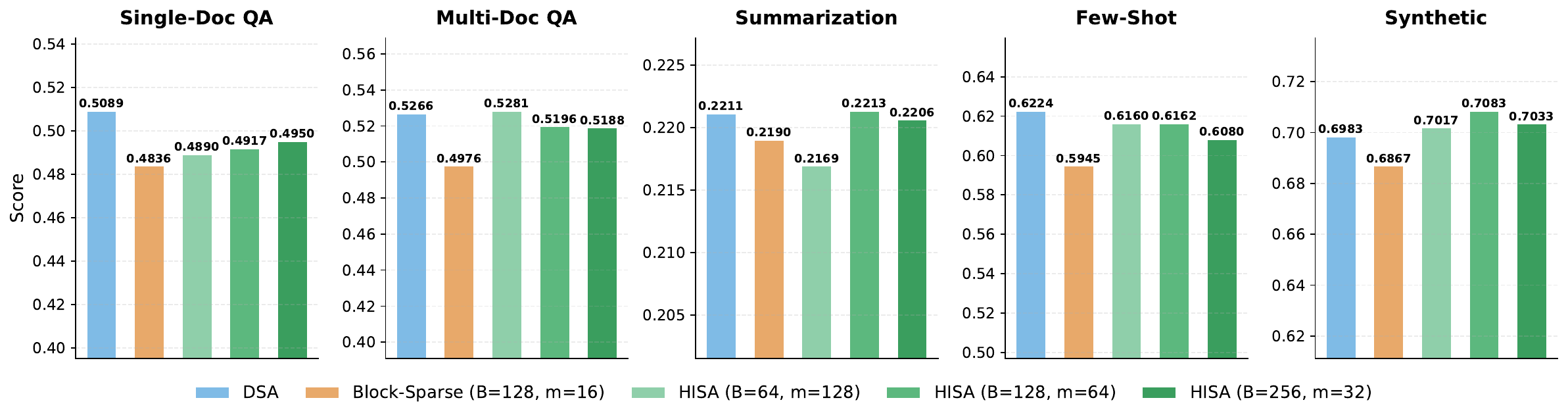}
    \caption{Ablation study on DeepSeek-V3.2.}
    \label{fig:hyperparams_dsv32}
\end{subfigure}
\begin{subfigure}[t]{\textwidth}
    \centering
    \includegraphics[width=\linewidth]{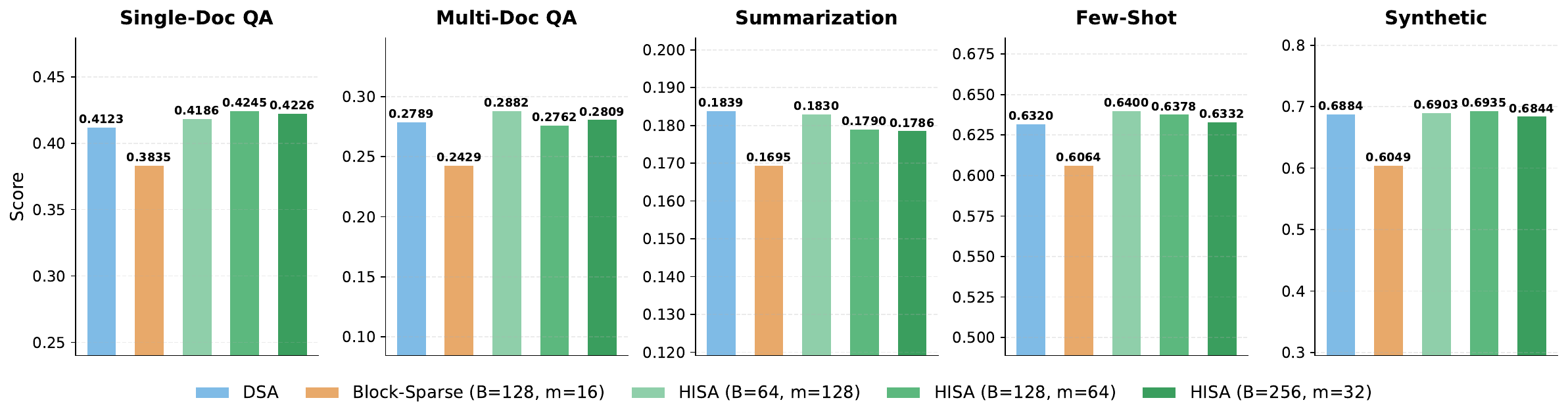}
    \caption{Ablation study on GLM-5.}
    \label{fig:token_indexer_glm5}
\end{subfigure}
\caption{LongBench scores under different indexer configurations. All three HISA variants use a candidate token pool of size $mB = 8192$ and a final token budget of $k{=}2048$, with different choices of block size $B$ and block-level top-$m$. The Block-Sparse baseline uses $B{=}128$ and $m{=}16$, corresponding to a candidate pool of 2048 tokens and no token-level refinement.}
\label{fig:hyperparams}
\end{figure}

Figure~\ref{fig:hyperparams} reveals two key findings. First, all three HISA configurations closely track DSA performance across all five task categories while consistently and substantially outperforming the Block-Sparse baseline. HISA is thus nearly lossless relative to the exhaustive flat scan, and token-level refinement is essential to this gap: even under the same block-level selection mechanism, the ability to prune low-relevance tokens \emph{within} selected blocks---which Block-Sparse cannot do---yields measurable quality gains.
Second, among the three HISA variants, the best block size is not uniform across sub-tasks and the trend in $B$ is not monotonic: $B{=}64$ and $B{=}128$ tend to edge out $B{=}256$ overall, but the ranking is not identical across all task categories, and there is no single HISA configuration that is optimal everywhere. We attribute this non-monotonicity to the current training-free mean-pooling aggregation rather than to the hierarchical indexing framework itself; we analyze the underlying mechanism in detail below.

\paragraph{Why the trend is non-monotonic.} Many of the per-task gaps in Figure~\ref{fig:hyperparams} are small and partly within inference/evaluation variance, so nearby configurations should not be over-interpreted; we instead focus on the two categories that show the clearest, opposite trends on DeepSeek-V3.2---MQA favors smaller $B$, while SQA favors larger $B$. A smaller $B$ is not uniformly better because, under training-free mean pooling, HISA's coarse filter carries two interacting error sources:
\begin{itemize}
    \item \textbf{Block-granularity error}: Stage~1 selects whole blocks rather than individual tokens, so smaller $B$ is, in principle, always better for this term alone, since finer block granularity is strictly closer to DSA's token-level selection. This error dominates for MQA, where evidence is fine-grained and scattered across multiple documents: under the fixed candidate budget $mB$, a larger $B$ forces a smaller $m$, so fewer, coarser blocks are retained and scattered relevant fragments are more likely to be missed, whereas a smaller $B$ with a correspondingly larger $m$ gives finer granularity and broader coverage.
    \item \textbf{Block-representation error}: each block is summarized by a single mean-pooled vector; because this aggregation is training-free, the summary is not a particularly strong one, so its effect on $B$ is not monotonic. For SQA, where the answer often lies in one long contiguous span, a moderately larger block can actually \emph{improve} accuracy by pooling the whole span into a single, easily-retrieved block vector---so accuracy first rises above the small-$B$ baseline, then turns over and declines once the block becomes so large that mean pooling blurs the useful span with irrelevant surrounding context.
\end{itemize}
We emphasize that the block-representation error is a consequence of training-free mean pooling, not of HISA's hierarchical framework itself. With a trained block-level summarizer, this error should shrink and the block-granularity error would dominate more uniformly, so smaller $B$ would more often improve accuracy as intuition suggests, restoring the more intuitive ``smaller $B$ is better'' trend; we expect such a setting to find a stable $B$ that matches DSA accuracy while preserving the hierarchical-indexing speedup (Section~\ref{sec:conclusion}). We verify the current trade-off directly by evaluating substantially larger block sizes (up to $B{=}2048$) under the same fixed candidate budget $mB{=}8192$; see Appendix~\ref{app:largeblock} for the full results and category-level breakdown. In the current training-free setting, we recommend $B{=}128, m{=}64$ as a practical default: it stays close to DSA across all task categories (Table~\ref{tab:longbench}) and underlies the end-to-end speedups reported in Section~\ref{sec:exp:e2e}, without requiring per-task hyperparameter search.

\subsection{Extended Block-Size Ablation}
\label{app:largeblock}

Appendix~\ref{app:hyperparams} attributes HISA's non-monotonic block-size trend to two interacting error sources under training-free mean pooling: a \emph{block-granularity error} that shrinks with smaller $B$, and a \emph{block-representation error} from mean pooling whose dependence on $B$ is not monotonic. To verify this trade-off directly, Table~\ref{tab:largeblock} extends the sweep to substantially larger blocks under the same fixed candidate budget $mB{=}8192$ on DeepSeek-V3.2---$(B{=}512, m{=}16)$, $(B{=}1024, m{=}8)$, and $(B{=}2048, m{=}4)$---alongside the $B{=}128, m{=}64$ default from Table~\ref{tab:longbench} for reference.

\begin{table}[h]
\centering
\caption{LongBench results for DeepSeek-V3.2 with substantially larger block sizes, at the same fixed candidate budget $mB{=}8192$ used in Figure~\ref{fig:hyperparams}. The $B{=}128, m{=}64$ row is the deployment default (reproduced from Table~\ref{tab:longbench} for reference); Avg.\ is computed over the same five task categories as Figure~\ref{fig:hyperparams} (excluding Code).}
\label{tab:largeblock}
\small
\begin{tabular}{lrrrrrr}
\toprule
\textbf{Config} & \textbf{SQA} & \textbf{MQA} & \textbf{Sum} & \textbf{FS} & \textbf{Syn} & \textbf{Avg.} \\
\midrule
$B{=}128, m{=}64$ (default) & 49.17 & 51.96 & 22.13 & 61.62 & 70.83 & 51.14 \\
$B{=}512, m{=}16$  & 49.75 & 51.78 & 22.06 & 60.63 & 69.83 & 50.81 \\
$B{=}1024, m{=}8$  & 49.66 & 51.07 & 22.04 & 60.45 & 69.83 & 50.61 \\
$B{=}2048, m{=}4$  & 48.65 & 50.49 & 22.00 & 61.39 & 68.83 & 50.27 \\
\bottomrule
\end{tabular}
\end{table}

The results confirm the two-error explanation from Appendix~\ref{app:hyperparams}. For MQA, where evidence is fine-grained and scattered across multiple documents, the block-granularity error dominates and score declines steadily as $B$ grows from 128 to 2048 (51.96 $\to$ 50.49). For SQA, where the answer often lies in one long contiguous span, the block-representation error dominates instead, and the trend is non-monotonic rather than flat: accuracy first \emph{rises} from $B{=}128$ to $B{=}512$ (49.17 $\to$ 49.75), as a moderately larger block more effectively pools the whole evidence span into a single, easily-retrieved block vector, then turns over and declines through $B{=}1024$ (49.66) and $B{=}2048$ (48.65) once the block becomes large enough that mean pooling blurs the useful span with irrelevant surrounding context.

\subsection{Failure Case Studies}
\label{app:failure-cases}

We analyze cases where DSA is correct while HISA is not, to understand when and why the coarse filter can miss the correct evidence. Mechanistically, HISA can only lose information relative to DSA when Stage~1 fails to retain the block containing the relevant tokens: once a block survives coarse filtering, Stage~2 applies the identical DSA token-level indexer within it. The following three representative cases, with exact model outputs, trace this loss back to the current training-free mean-pooling instantiation: mean pooling can blur fine-grained entity relations, short local causal-action spans, or an isolated needle diluted by a much larger surrounding haystack, causing Stage~1 to assign a low score to the block that actually contains the answer.

\paragraph{Entity-relation blurring.} A representative case from the LongBench 2WikiMQA subset asks: ``When did William Le Poer Trench's father die?'' DSA answers correctly, ``26 April 1872,'' while HISA outputs a distractor date, ``24 November 1837.'' Answering requires first linking ``William Le Poer Trench'' to his father, ``William Trench, 3rd Earl of Clancarty,'' and then retrieving the father's death date, while the surrounding context also names several other family members with similar titles. When the mean-pooled block representation blurs multiple overlapping entity relations together, the block containing the correct death date can receive a low Stage-1 score and never reach Stage~2.

\paragraph{Local causal-action dilution.} A second case, from NarrativeQA, asks how slime gets into Dana's apartment; the decisive evidence is a short local action span in which water from the faucet turns to slime and settles in the tub. DSA gives the precise answer, ``through the pipes into the bathtub,'' while HISA gives the more generic ``through the apartment's plumbing.'' A short, relation-bearing action span can be outweighed at Stage~1 by nearby blocks that contain more salient but less precise lexical matches (e.g., ``slime,'' ``bathtub,'' ``apartment''), biasing block selection away from the exact causal evidence.

\paragraph{Needle dilution.} A third pattern is specific to Needle-in-a-Haystack: the relevant evidence there is an isolated needle whose signal occupies only a few tokens. When such a needle is pooled together with a much larger surrounding haystack, its signal can be dominated by the haystack in the mean-pooled block vector, so the block containing the needle may receive a low Stage-1 score.

All three patterns trace back to the same root cause: the current training-free mean pooling, not the hierarchical indexing framework itself---once a relevant block is retained, token-level accuracy is unaffected. Learned block-level representations in a trained HISA variant may better preserve such evidence while retaining the two-stage efficiency benefit, which we leave to future work (Section~\ref{sec:conclusion}).

\clearpage
\section{Phase-Sensitivity Experiment Details}
\label{app:hisa-phase-details}

\paragraph{Inputs and reproducibility.}
We use the data construction and independent validation code released by
\citet{zhu2026periodicweakspots}, pinned to commit
\href{https://github.com/UltimateJupiter/Periodic-Weak-Spots/tree/09c109910ba29b834f54441d96b442ab9e1634ae}{\texttt{09c1099}}.
Each prompt contains 16,000 four-token key--value records, with target record index 248 (zero-based), a mean target position of approximately 64K, and a total encoded text length of 128,000 tokens.
There are 256 bindings, each instantiated at all eight source-position phases, yielding 256 observations per phase and 2,048 per model/variant.
The three GLM arms share exactly the same sample IDs, bindings, answers, and inputs.
This is a behavioral reproduction: our public Paul Graham filler corpus does not have the paper's reported byte hash, and our DeepSeek-V4.1 evaluation uses 256 rather than the paper's 2,560 observations per phase.

GLM uses its own tokenizer. To obtain enough context-stable, disjoint single-token keys and values, we quote the query key at its two occurrences; the resulting query has 35 tokens, and the trailing filler length is adjusted accordingly.
The record geometry, phase-balancing design, and independent structural checks are otherwise inherited from the released generator.
All 2,048 GLM prompts pass validation, and one prompt per phase is additionally checked against Hugging Face tokenization.
GLM phases are the actual token positions of the source-record starts modulo eight.
For DeepSeek, we retain the source-phase labels supplied by the authors' dataset; any tokenizer-added leading-special-token offset is recorded separately in provenance.
Inputs therefore need not be identical across model families.

\paragraph{Model and selection details.}
The checkpoints are DeepSeek-V4-Flash-0731, DeepSeek-V4.1-Flash, and GLM-5.3-Flash; the first is not the Flash-Base checkpoint.
GLM-5.3-Flash combines 11 sparse-attention layers with 34 linear-attention layers.
Our interventions modify selection only in the 11 sparse layers; the linear-attention layers and all model weights remain unchanged.
Native selection retains 512 complete four-token blocks and the native unfinished-block tail.
HISA retains 2,048 complete blocks plus up to three tail candidates, and applies token-level refinement to obtain at most 2,048 causal token indices.
DSA directly selects at most 2,048 tokens from all causal history.
The original indexer Q/K projections, head weights, positional transformations, and FP8/Hadamard conventions are retained.
Both interventions share the native FP8 MQA token-scoring kernel.
They are inference-time interventions on a checkpoint trained with its native block indexer, not separately trained DSA/HISA models; any adaptation disadvantage is not quantified by this experiment.

\paragraph{Runtime control.}
Historical full runs use vLLM 0.31.0, PyTorch 2.13.0, Transformers 5.17.0, eight-way tensor parallelism with expert parallelism, and eight NVIDIA L20Z GPUs.
We use eager execution, one sequence at a time, no prefix caching, 8,192-token prefill chunks, seed 33377335, and greedy generation of at most eight tokens from raw text.
HISA and DSA are evaluated in paired order in the same engine; the native control is rerun separately with matching experimental-buffer allocation and the same frozen FlashInfer tactic settings.
Worker audits verify that native selection remains unmodified in this control.
The earlier native run is retained for provenance but is not the plotted reference.
The HISA diagnostic implementation computes additional token scores and masks out positions outside the candidate set; it is not a candidate-only optimized kernel, so its runtime cannot establish a deployment speedup.

\paragraph{Scoring.}
The original metric tests whether the whitespace-trimmed completion starts with the gold answer.
For GLM's default quote-tolerant metric, we first remove leading whitespace, then one contiguous prefix drawn from a fixed set of quote characters, then leading whitespace again, before testing the gold-answer prefix.
The quote set is U+0022, U+0027, U+0060, U+201C, U+2018, and U+00AB.
These operations are applied once in this order; they are not repeated to remove arbitrarily interleaved quotes and spaces.
We do not search for the answer elsewhere in the completion or apply case folding.
As in the original metric, continuation after a correct prefix does not affect correctness.
The quote-tolerant metric was introduced after observing formatting failures, rather than preregistered, and no samples were reselected on this basis.

\begin{table}[htbp]
\centering
\small
\caption{GLM-5.3-Flash results under both scoring rules. Accuracy is in percent; $\Delta_8$ is the maximum minus minimum of the eight phase accuracies, in percentage points.}
\label{tab:hisa-phase-both-metrics}
\begin{tabular}{lrrrr}
\toprule
& \multicolumn{2}{c}{Quote-tolerant} & \multicolumn{2}{c}{Original} \\
Method & Accuracy & $\Delta_8$ & Accuracy & $\Delta_8$ \\
\midrule
block-indexer (native) & 84.03 & 5.86 & 76.81 & 7.03 \\
HISA & 86.33 & 1.95 & 73.19 & 3.12 \\
DSA & 86.38 & 2.73 & 76.22 & 4.30 \\
\bottomrule
\end{tabular}
\end{table}

In particular, HISA's original-metric accuracy decreases from 76.81\% to 73.19\%, whereas its quote-tolerant accuracy increases from 84.03\% to 86.33\%.
The improvement is therefore metric-dependent and must not be described as an unconditional gain in answer-prefix accuracy.
The original-metric DSA accuracy is 76.22\%.

\paragraph{Uncertainty and scope.}
We jointly resample the same 256 binding IDs across the three arms, preserving all eight phases within each binding (10,000 bootstrap replicates; NumPy seed 173021).
For each replicate, we recompute $\Delta_8$ and subtract the intervention gap from the native-control gap; the reported 95\% intervals are percentile intervals of this paired difference.
Both quote-tolerant gap-reduction intervals include zero.
The study therefore supports a descriptive reduction of the observed phase variation, not statistical equivalence across phases, complete removal of phase sensitivity, or a guarantee across tasks, contexts, seeds, and models.
The interventions jointly change final selection granularity and, for HISA, candidate coverage; the experiment does not separately identify the contributions of these changes or exclude other sources of positional sensitivity.

\paragraph{Artifacts.}
Code, exact scoring rules, per-example completions, and provenance are available at
\href{https://github.com/MuLabPKU/TransArch/tree/815ea014f6001ad05f04c4715b0d5e4e7134e9e1/OpenPWS}{OpenPWS (commit \texttt{815ea01})}.
The plotted results are the archived full runs, not a new full evaluation of the compact release code.
That refactoring was checked on eight prompts across all three GLM modes: all 24 outcomes agree with the historical records under both metrics; native token IDs agree for six of eight prompts, while HISA/DSA agree for all eight each.
These checks establish limited implementation validation rather than bitwise equivalence of full runs.

\end{document}